\documentclass[letterpaper, 10 pt, conference]{ieeeconf}  % Comment this line out
\IEEEoverridecommandlockouts                              % This command is only
\usepackage[utf8]{inputenc}
\usepackage[T1]{fontenc}
\usepackage{acronym}
\usepackage{graphicx}
\usepackage{multirow}
\usepackage{microtype}
\usepackage{makecell}
\usepackage{subcaption,siunitx,booktabs} 
\usepackage{authblk}
\usepackage{times}
\usepackage{epsfig}
\usepackage{amsmath}
\usepackage{amssymb}
\title{\LARGE \bf
Manipulating Medical Image Translation with Manifold Disentanglement}

\author{
    Siyu~Liu$^{1}$,
    Jason~A.~Dowling$^{2}$
    Craig~Engstrom$^{1}$,
    Peter~B.~Greer$^{3}$,
    Stuart~Crozier$^{1}$,~and~
    Shekhar~S.~Chandra
}

\affil[$^{1}$]{School of Information Technology and Electrical Engineering, University of Queensland, Australia}
\affil[$^{2}$]{The Australian e-Health Research Centre, CSIRO, Australia}
\affil[$^{3}$]{School of Mathematical and Physical Sciences, The University of Newcastle, Australia}       

\begin{document}
\maketitle
\thispagestyle{empty}
\pagestyle{empty}
\begin{abstract}
    Medical image translation (e.g. CT to MR) is a challenging task as it requires I) faithful translation of domain-invariant features (e.g. shape information of anatomical structures) and II) realistic synthesis of target-domain features (e.g. tissue appearance in MR). In this work, we propose \ac{MDGAN}, a novel image translation framework that explicitly models these two types of features. It employs a fully convolutional generator to model domain-invariant features, and it uses style codes to separately model target-domain features as a manifold. This design aims to explicitly disentangle domain-invariant features and domain-specific features while gaining individual control of both. The image translation process is formulated as a stylisation task, where the input is ``stylised" (translated) into diverse target-domain images based on style codes sampled from the learnt manifold. We test \ac{MDGAN} for multi-modal medical image translation, where we create two domain-specific manifold clusters on the manifold to translate segmentation maps into pseudo-CT and pseudo-MR images, respectively. We show that by traversing a path across the MR manifold cluster, the target output can be manipulated while still retaining the shape information from the input.
\end{abstract}

\section{Introduction}
\ac{GAN}~\cite{gan} and conditional \ac{GAN}~\cite{cgan} have been rising in popularity for medical image synthesis. Conditional \ac{GAN} is currently the dominant method for cross-modality medical image translation. For example, in MR-only radiotherapy treatment~\cite{Dowling2015}, a conditional \ac{GAN} can be used to ``retrieve" missing CT images from other available imaging modalities. The generative aspect of conditional \ac{GAN} can also be useful in medical imaging analysis research, for example, a robust \ac{GAN} capable of generating realistic and diverse examples can be used as a data augmentation tool to improve the performance of other models~\cite{gan-aug,gan-aug2}.

A \ac{GAN}-based image translation framework consists of a mapping function $G$ from the source domain $A$ to the target domain $B$ and $B \approx G(A)$. In medical image translation, since different imaging domains (modalities) contain mutual as well as exclusive features, it is paramount that $G$ learns to preserve important domain-invariant features (e.g. anatomical structures and shape information). At the same time, it also needs to learn diverse features (e.g. tissue appearance and image contrast) specific to domain $B$ such that $G(A)$ visually resembles $B$. In most image translation \acp{GAN}, these two types of features are intertwined and cannot be individually controlled. In medical image translation, we argue it is desirable to disentangle domain-invariant features and domain-specific features such that they can be manipulated individually. For example, In a CT-MR translation task, we may want to diversify features specific to the MR domain (such as contrast and tissue appearance) while leaving the underlying anatomical structures intact. We may also want to alter the target-domain of the output (e.g. PET instead of MR) with the same constraint.

\begin{figure}[t]
  \includegraphics[width=\linewidth]{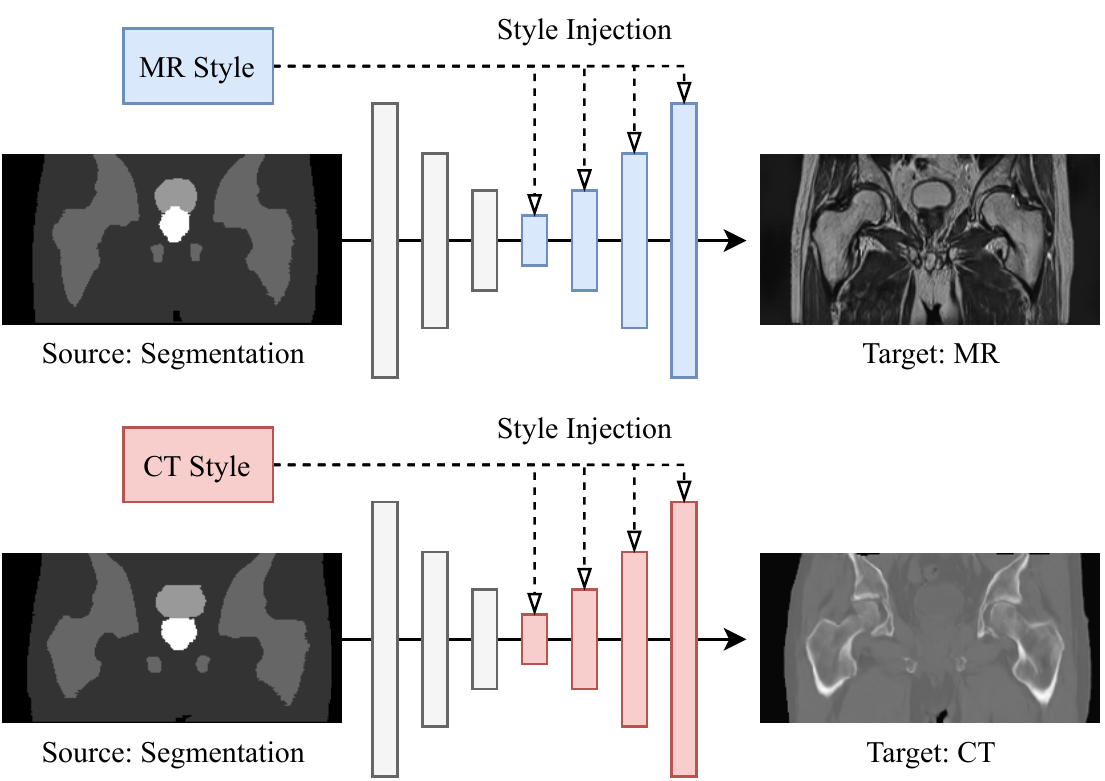}
  \caption{Proposed manifold disentanglement \ac{GAN} for (style-based, multi-modal) medical image domain translation. The single generator supports multiple modalities as styles across domains.}
  \label{fig:domain-trans}
\end{figure}
The StyleGAN~\cite{stylegan, stylegan2} framework is a natural candidate for our objective. A StyleGAN generator relies on style code injection to manipulate the output, and the style codes essentially form a manifold that controls the output. We can logically formulate our objective as a stylisation problem. As in Figure~\ref{fig:domain-trans}, a shared encoder-decoder generator network is used to extract and retain domain-invariant features. At the same time, we learn a disentangled manifold which provides style codes to “stylise” (translate) the input to the correct target-domain. Based on this idea, we propose \ac{MDGAN}, a powerful style-based generative framework for medical image translation. The contribution of this framework can be summarised as

\begin{itemize}
    \item We harness the StyleGAN framework to explicitly disentangle domain-invariant features and domain-specific features. The generator implicitly learns a manifold of target-domain features for image translation. The style codes are manifold clusters embedded on this manifold. This manifold provides control of the translated images, but domain-invariant features from the source input are still faithfully retained as a result of feature disentanglement.
    
    \item The generator is trained to interpret and generate multi-modal images based on multiple manifold clusters. This property enables multi-modal medical image translation with a shared generator and separately learnt manifold clusters. In our case,  \ac{MDGAN} learns a CT manifold and a MR manifold to generate pseudo-CT and pseudo-MR images from input segmentation maps.
    
    \item By sampling the manifold, we can explore and interpolate the latent space to generate diverse images. All the images are different in appearance, but without violating domain-invariant features from the source input.
\end{itemize}

% \ac{MDGAN} was tested for multi-modal medical image domain translation from segmentation maps to pseudo-MR and pseudo-CT. 
We use a shared-generator to pass domain-invariant information, and inject style codes from two separate manifold cluster networks to synthesise realistic MR and CT images. The proposed framework also goes beyond one-to-one domain mapping and can produce diverse outputs for a given single input segmentation. Finally, we perform dimensionality reduction on the style codes to reveal two well formed manifold clusters. By sampling style codes along geodesic paths across the MR manifold cluster, we observed smooth and systematic transitions in tissue appearances, while keeping the shape information of the anatomical structures consistent with the input segmentation.

\section{Related Work}
\subsection{Generative Adversarial Networks}
The vanilla \ac{GAN}~\cite{gan} is composed of a generator network and a discriminator network. During training, the generator's performance is improved by competing against the discriminator, which is given the opposite objective. The min-max objective of a basic \ac{GAN} is defined as 
\begin{equation}
\begin{gathered}
      \underset{G}{\textnormal{min}}~\underset{D}{\textnormal{max}}~V(D, G) = \mathbb{E}_{x\sim p_{data(x)}} [\textnormal{log}D(x)]\\
    ~+~\mathbb{E}_{z\sim p_{z(z)}}[\textnormal{log}(1-D(G(z))]  
\end{gathered}
\end{equation}

The generator and the discriminator achieve optimal performance when the \ac{GAN} reaches a state of Nash Equilibrium. In practice, the training is often unstable and characterised by failures such as mode collapse, non-convergence and vanishing gradient. Efforts have been made to stabilise the training process of \ac{GAN}. While notable work, including LSGAN~\cite{lsgan}, WGAN~\cite{wgan}, WGAN-GP~\cite{wgan-gp} and PGGAN~\cite{pggan} have proposed novel methods to alleviate these failures modes and improve performance. Training a \ac{GAN} is still very much an empirical process, and the training procedures can be highly domain sensitive.

% \subsection{Conditional Generative Adversarial Networks}
Conditional \ac{GAN}~\cite{cgan} is a type of \ac{GAN} that involves data synthesis based on some conditional input. Usually, the input to the generator is highly correlated to the output and can be exploited to gain control of the generated data. Conditional \acp{GAN} are best known for their success in image domain translation. For example, from sketches to photo-realistic images of the sketched objects~\cite{pix2pix}. These networks often rely on auxiliary losses such as L1, L2 and perceptual losses~\cite{perceptual-loss} to enforce the correlation and consistency across the source and the target domains. 

\subsection{Style-based \ac{GAN}}
Style-based \ac{GAN}~\cite{stylegan} stands on its own as an alternative approach to unconditional image synthesis. It rethinks image synthesis from the perspective of style transfer. Huang et al.~\cite{adain} explored the profound effect of activation normalisation in \ac{CNN} and proposed \ac{AdaIN} as a means of real-time style transfer. In comparison to traditional gradient-based style transfer methods~\cite{neural-style-transfer}, \ac{AdaIN} exhibits superior versatility and control of the output style. StyleGAN~\cite{stylegan} was the first to harness the power of \ac{AdaIN} in a generative adversarial framework. Instead of starting with a latent noise vector, the StyleGAN generator applies \ac{AdaIN} at various points of the network to inject a learnt latent (style) vector. This alternative approach to image generation achieves unprecedented control of the generated image at all scales (from global structure to local details). Additionally, the network uses a progressively growing training scheme, mini-batch standard deviation~\cite{pggan}, path length regularisation, style-mixing regularisation to optimise performance and stabilise training. The successor to StyleGAN, StyleGAN2~\cite{stylegan2}, was published with various significant refinements. First, \ac{AdaIN} was removed in favour of modulated convolution to alleviate normalisation artefacts in the output. Second, the progressively growing networks were simplified to residual architectures for easier training. Lastly, a new path length regularisation was used to improve image quality and network invertibility.  The main limitation of the original StyleGAN and StyleGAN2 is their unconditional nature, which is unsuitable for image translation tasks. Recently, Pixel2Style2Pixel~\cite{encode-style} proposed an image encoder network as an extension to the StyleGAN framework. The encoder network maps the inputs to codes, which enables domain translation using the StyleGAN framework. Another prominent conditional extension to StyleGAN is StarGAN~\cite{stargan2}, which employs \ac{AdaIN} and cycle consistency to achieve unpaired multi-domain translation.
 
\subsection{Generative Adversarial Networks in Medical Imaging Analysis}
Both conditional \ac{GAN} and unconditional \ac{GAN} have been widely adopted for medical imaging analysis~\cite{review}. Medical image synthesis using \ac{GAN} has been shown effective for data augmentation, which may enhance existing deep learning models facing data scarcity. For example, \cite{gan-aug} uses \ac{GAN}-generated CT scans to improve segmentation accuracy, and~\cite{gan-aug2} uses three generative networks to synthesise three types of lesions to improve classification accuracy. Conditional \ac{GAN} is useful for a multitude of applications beyond data augmentation thanks to their versatility. The most common application of conditional \ac{GAN} is image domain translation~\cite{medgan}. For example, segmentation map to medical image~\cite{seg-img, dual-gan}, MR reconstruction~\cite{mr-recon, cs1}, image denoising~\cite{denoise, denoise-percept} and cross-modality translation~\cite{med-trans, mr-ct, ct-mr}. Many of these methods are based on the popular Pix2Pix framework, which relies on paired data across domains. When pair-wise labels are not available, CycleGAN~\cite{cyclegan} and UNIT~\cite{unit} are used for semi-supervised learning on unpaired images~\cite{ med-cycle, med-cycle2, cycle-med3, susan, cyc-unit}. 

\subsection{\ac{MDGAN}}
Compared to conventional \ac{GAN}, the StyleGAN framework introduces a more profound approach to manipulate outputs using an external style input. As described in the Introduction, it fits naturally within our objective of creating a versatile medical image translation framework based on feature disentanglement. However, the original StyleGAN framework is fundamentally unconditional, and its manifold does not provide disentanglement of domain-related features. While Pixel2Style2Pixel is one step closer to our objective due to the addition of a conditional input and its diverse outputs, it does not provide explicit disentangled of domain-invariant features and domain-specific manifold for independent control. At the same time, it is not designed for multi-modal applications. StarGAN has also been considered a candidate for our task as it is a powerful multi-domain image translation network. However, it does not learn a manifold of domain-specific features. Comparing to other existing approaches for medical image translation, most of them are not multi-modal, thus requiring a designated network for each target domain. An immediate advantage of the proposed stylisation approach is the sharing of domain-invariant knowledge, only a small manifold cluster is learnt for each domain, and the entire generator is shared. Most of the methods in the medical imaging context only perform one-to-one mappings on a given input, which does not capture the true dynamics of image translation tasks. For example, one segmentation map can be theoretically mapped to infinitely many valid target images. We are also not aware of other work that formulates multi-domain medical image translation as a general stylisation problem with disentangled manifolds. The closest use of a style-based \ac{GAN} in medical imaging analysis is \cite{latent-manipulation}, which uses the original StyleGAN to explore the latent space of medical images. We are also aware that unpaid image translation enabled by CycleGAN and UNIT can be more desirable for some applications, but to the best of our knowledge, there has not been method, especially in medical image analysis, based on exploitable manifolds of disentangled features. We consider a cycle version of this framework a potential future extension.

\begin{figure*}[hbt]
    \centering
    \includegraphics[width=.9\linewidth]{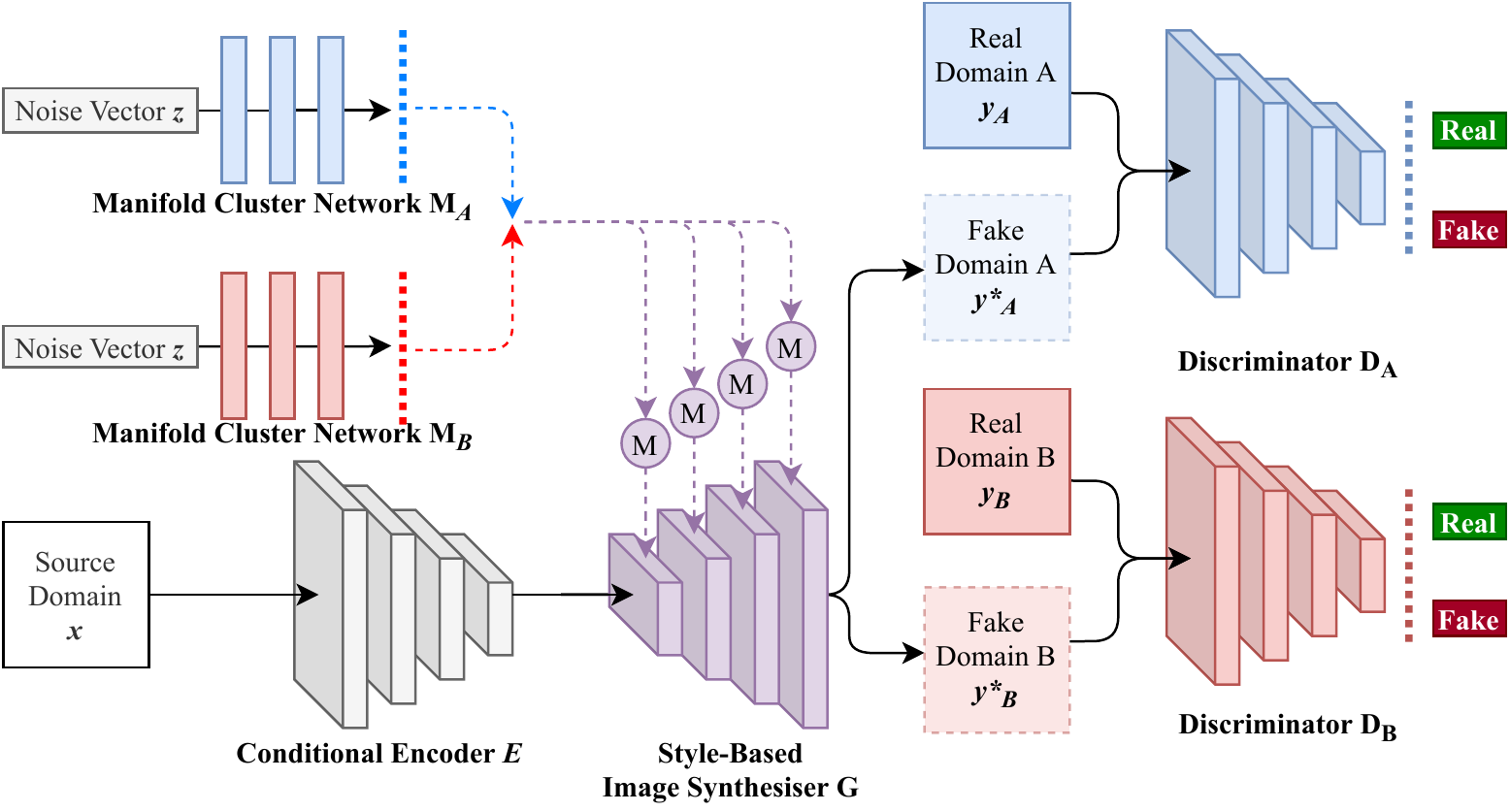}
    \caption{Network architectures for the proposed framework. The generative networks $E$ and $G$ are shared. $G$ uses modulated convolution (denoted $M$) for stylisation based on some external style code. The multiple style code manifold networks represent multiple manifold clusters on the overall manifold.}
    \label{fig:generator}
\end{figure*}

\section{Methods}
The objective of the proposed method is to achieve disentangled representations of domain-invariant features and domain-specific features, where the domain-invariant features are translated to the target domain according to style codes sampled from learnt manifold clusters. In this section, we formulate a framework for one-to-many medical image translation using this idea.

\subsection{Proposed Framework}
\label{sec:framework}
Improving upon the foundation of StyleGAN, the proposed \ac{MDGAN} consists of four networks: conditional encoder ($E$), style-based image synthesiser ($G$), manifold cluster network ($M$) and discriminator ($D$). $E$ and $G$ are shared networks which encapsulate domain-invariant information (such as anatomical structure and shape information). For each target domain $T$, a corresponding manifold cluster network ($M_T$) and discriminator ($D_T$) are trained to encourage the formation of manifold clusters.

The sub-network interactions are illustrated in Figure~\ref{fig:generator}. Given a source domain input $x$ and a desired target domain $T$, $E$ produces a shared latent representation of the input $w = E(x)$ which learns domain-invariant features, $G$ is supplied both $w$ and a domain-specific style code $s_T = M_T(z_{\textnormal{noise}}\sim N(0, 1))$ as conditional inputs to synthesise the translated image $y^*_T = G(w,\ s_T)$. Note that $w$ and $s_T$ are deliberately separated to disentangle shared features and domain-specific features.  The role of $s_T$ is to modulate the convolutional weights in $G$ to achieve the desired output style. Compared to a random noise vector, $s_T$ is more interpretable to $G$ as it is fundamentally a manifold cluster of the features specific to domain $T$. Finally, the discriminator for each target domain is a binary classifier $D_T$ which aims to distinguish real data $y_T$ from the fake $y^*_T$.

For each training step, we randomly sample a target domain $t\in T$ and train $M_t$, $D_t$ with the shared image synthesiser $G$ and conditional encoder $E$.
% IMPORTANT
% alternating CA CB
% how was the latent mapping network trained
% explain the perceptual loss a bit more symbol
\subsection{Network Architecture Details}
$M_T$ is a fully-connected network with four 384-unit hidden layers. It learns a mapping from a 384-d noise vector $z\sim N(0, 1)$ to a 384-d intermediate manifold cluster $w$. $E$ is a fully-convolutional network and $D$ is a re-implemented StyleGAN2~\cite{stylegan2} with conditional inputs. $E$ contains four convolutional blocks (16, 16, 32 and 64 filters) which progressively down-sample the input while expanding the feature depths. $G$ contains convolutional blocks of depths 256, 128, 64 and 48, and its structure mirrors that of $E$ to recover the original scale of the input. Residual connections~\cite{stylegan2} are used in both $E$ and $G$ to improve the connectivity between neighbouring blocks. Like the original StyleGAN, we also incorporate noise feature maps in $G$ to introduce fine-grained variations. $D_T$ uses a similar structure to $E$ but the filter depths are increased to 48, 64, 128 and 256. The final feature maps of $D_T$ are mapped to a confidence score using a densely connected layer.

All of the convolutional layers in $G$ are modulated convolution as used in StyleGAN2. As below, modulated convolution performs weight re-normalisation based on some affine-transformed external style vector. In the proposed framework, this re-normalisation procedure provides the mechanism for freely switching among multiple target domains as well as produce diverse outputs.

\begin{equation*}
    w'_{ijk} = (s_i \cdot w_{ijk}) / \sqrt{\sum_{i, k}(s_i \cdot w_{ijk})^2 + \epsilon}
\end{equation*}

Like the modulated convolution in StyleGAN2, $s_i$ comes from an external style input. $i, j$ and $k$ enumerates the input feature maps, output feature maps and spatial dimensions respectively.

\subsection{Proposed Training Procedure}

\begin{figure}[hbt]
    \centering
    \includegraphics[width=.9\linewidth]{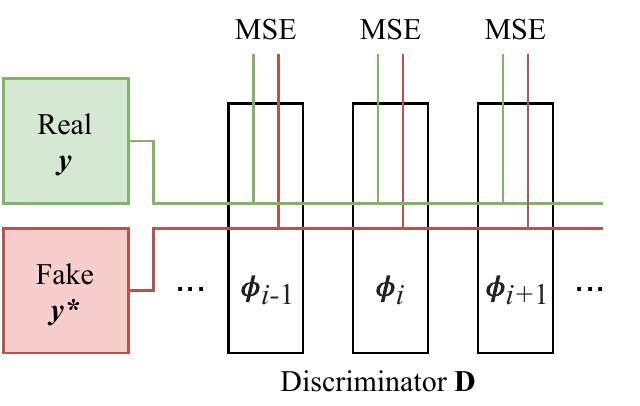}
    \caption{The perceptual loss uses the intermediate outputs of $D_T$. $\phi_i$ stands for the output of the $i$-th convolutional layer and $i \geq 3.$}
    \label{fig:perceptual}
\end{figure}

The primary losses of the proposed framework include a non-saturating adversarial loss $L_{\textnormal{adv}}$~\cite{gan}, a perceptual reconstruction loss $L_{\textnormal{rec}}$ and an $R_1$ gradient penalty $L_{\textnormal{gp}}$~\cite{convergence} term:
\begin{multline*}
    L_{\textnormal{adv}} = \mathbb{E}[\textnormal{log}D_T(y_T)]~-~\mathbb{E}[\textnormal{log}(D_T(G(E(x), M_T(z))]\\
    L_{\textnormal{rec}} = \sum^{n}_{i = 3}||\phi_i(G(E(x), M_T(z))) - \phi_i(y_T))||^2\\ 
    L_{\textnormal{gp}} = \mathbb{E}[||\nabla D_T(y_T)||^2]\\
\end{multline*}

When computing $L_{\textnormal{rec}}$, the perceptual network for target domain $T$ is its corresponding discriminator $D_T$. As Figure~\ref{fig:perceptual}, $L_{\textnormal{rec}}$ captures the perceptual difference based on the intermediate outputs (denoted $\phi$) of $D_T$ from the third convolution layer onward.

Most \ac{GAN} frameworks used for image translation are one-to-one mapping networks, which arguably resembles the undesirable effect of mode collapse. With the proposed framework, we can achieve one-to-many image translation by ensuring the output is diverse and valid. This is done by imposing an additional diversification regulariser (similar to~\cite{divsersify}) as below to ensure the output is well-conditioned on the style codes. Our experiments show that the model tends to collapse and produce similar outputs without this regulariser.
\begin{equation*}
    L_{\textnormal{div}} = -||G(E(x), M_T(z_1)) - G(E(x), M_T(z_2))||
\end{equation*}

The total loss is finally defined as follows. $\mu$ and $\lambda$ are scaling factors for the reconstruction loss and gradient penalty term, respectively. $\lambda_{div}$ is the scaling factor for the diversification loss. We use $\lambda_{div} = 1$, and unlike StarGAN2~\cite{stargan2}, we avoid using decay as it results in mode collapse in our case.
\begin{equation*}
    L_{\textnormal{total}} = L_{\textnormal{adv}} + \mu L_{\textnormal{rec}} + \lambda L_{\textnormal{gp}} + L_{\textnormal{div}} * \lambda_{div}
\end{equation*}

For training, we use Adam~\cite{adam} optimiser with a learning rate of 0.0001 for $G$ and $D$, and a learning rate of 0.000001 for all the other networks. The models are trained for 72 hours on an NVIDIA P100 GPU, which is equipped with 16GB of VRAM to process a batch of 8 images at a time.

\subsection{Experiment}
We test the proposed framework by performing domain translation from segmentation to MR and CT scans. The dataset is a manually segmented 3D prostate dataset with 211 MR and 42 CT scans. The scans are collected in a prostate cancer treatment study of 42 patients over the course of 8 weeks~\cite{Dowling2015}. Each 3D image is $256\times 256 \times 128$ in resolution and is manually labelled with five foreground classes: body, bone, bladder, rectum and prostate. During training, we randomly sample $ 256 \times 128$ images from the centre 40 slices of the coronal plane. All the input images are prepossessed by mapping their pixel intensity ranges to [0, 1]. To critically test the capability of \ac{MDGAN}, we deliberately avoid any data augmentation in the training process.

Our framework takes the segmentation maps (one-hot encoded) as the input, and $G$ stylises them to arrive at the target domains $T = \{CT, MR\}$. As described in \ref{sec:framework}, the generative part of the framework only requires two separately learnt manifold clusters networks $M_{CT}$ and $M_{MR}$, which are inexpensive to train. The two expensive components $E$ and $G$ are fully shared. $D_{CT}$ and $D_{MR}$ are also separately trained networks, but they do not contribute to inference and can be discarded after training.

Finally, we perform dimensionality reduction on the style codes to explore the learnt manifold of each domain. This is done by sampling 10,000 styles codes from $M_{MR}$ and $M_{CT}$ each, and map them to 2D space using UMAP~\cite{umap} (minimum distance of 0.2 and 5 neighbours). Manifold interpolation was performed on the MR style codes (instead of CT because MR scans contain more complex features) to observe visual transitions in the MR domain. For a given segmentation input, a geodesic path with 36 points across the manifold is selected, and 36 images are generated.

\begin{figure*}[t!]
    \centering
    \includegraphics[width=.9\linewidth]{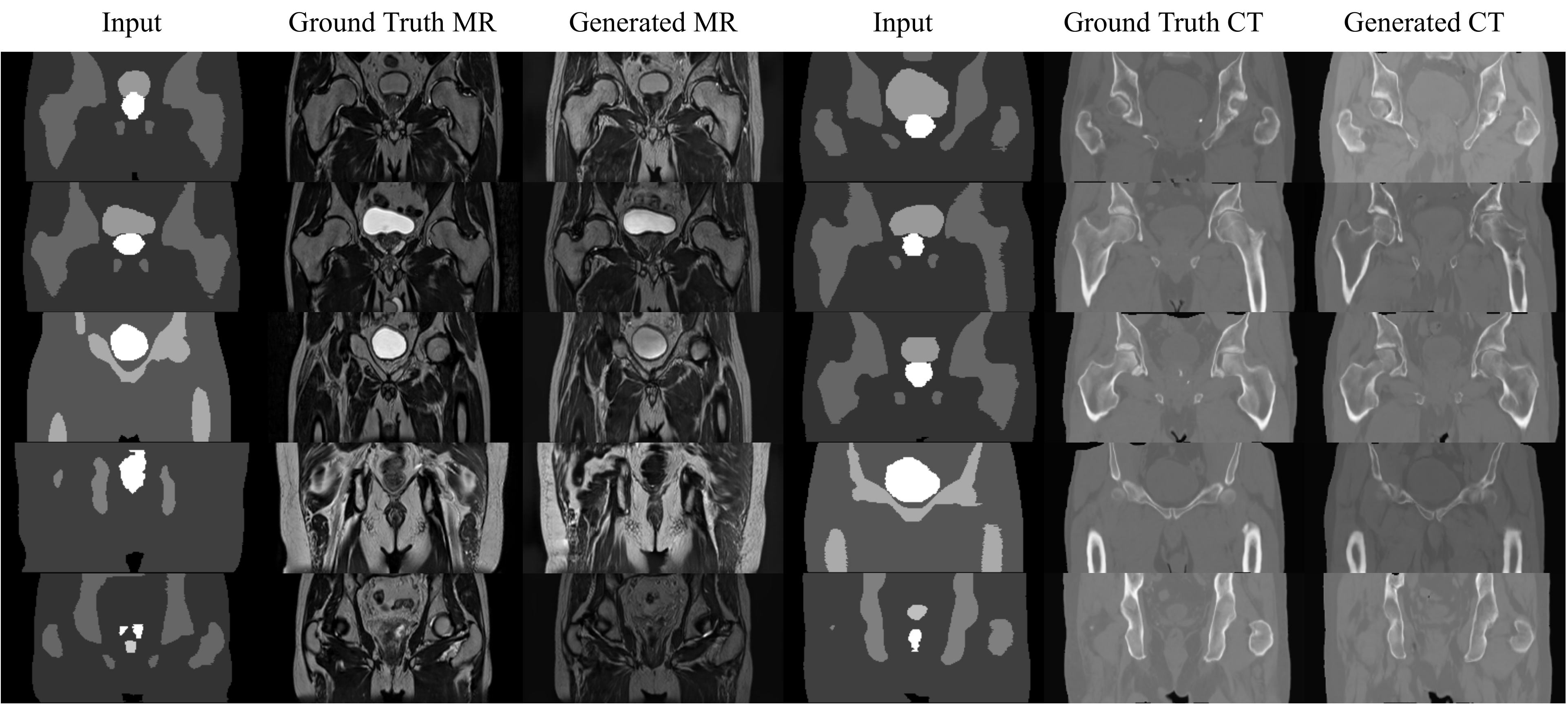}
    \caption{Representative results from the proposed \ac{MDGAN} framework for pseudo-CT and pseudo-MR generation.}
    \label{fig:results-viz}
\end{figure*}

\section{Results} 
In this section, we present the results from the medical image domain translation task, where our analysis will focus mainly on the generative capabilities, as well as exploring the inner workings of the disentangled manifold. Further results on other datasets are explored and provided in supplementary materials.

\subsection{Generator Results}
Figure~\ref{fig:results-viz} presents representative results generated using \ac{MDGAN}. All the generated images are acquired using a shared instance of the proposed style-based generator. It can be seen that the MR and CT outputs retain consistent shape information from the input segmentation maps. We also tested the robustness of the proposed model by elastically deforming the input segmentation and sampling a large number of different noise inputs. No noticeable failure cases were observed. This can be attributed to the learnt manifold clusters, which map random noise inputs to a more ``interpretable" latent space~\cite{stylegan} thus avoiding invalid combinations of features.

\subsection{Quantitative Results}
\begin{figure}[t!]
    \centering
    \includegraphics[width=0.7\linewidth]{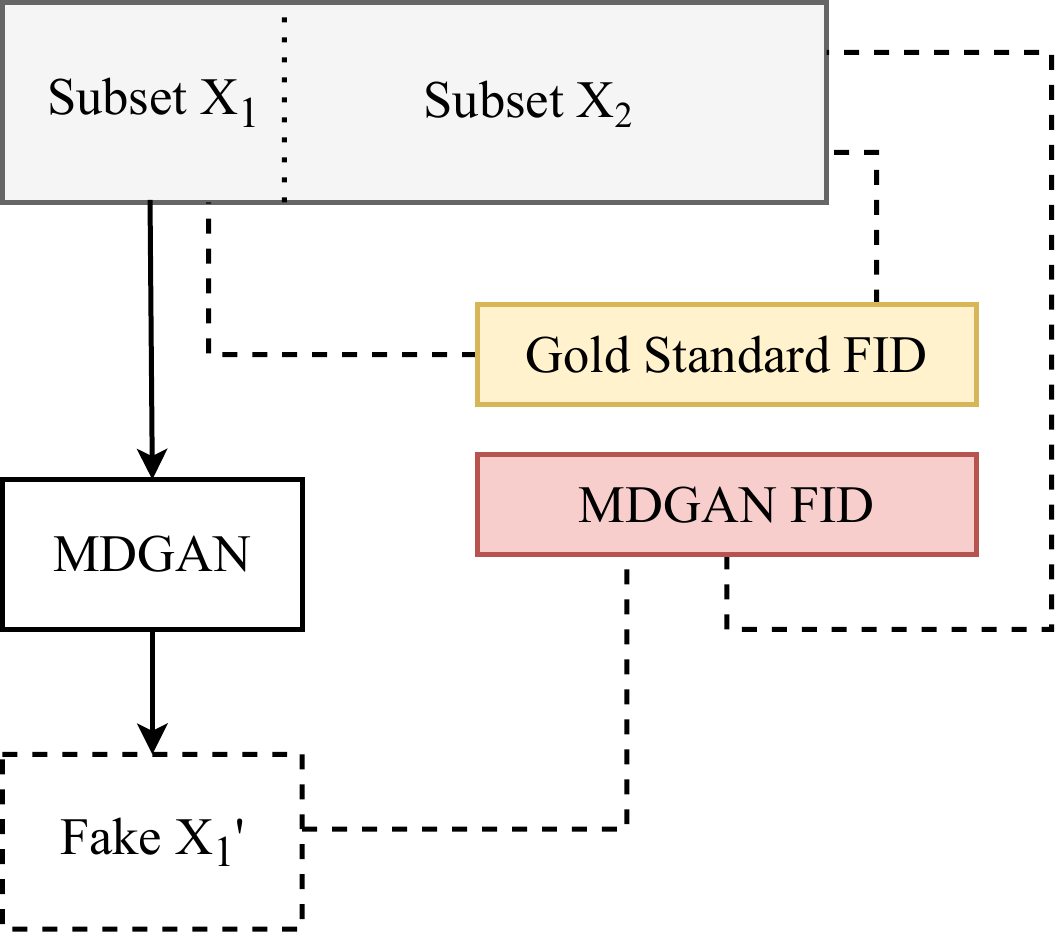}
    \caption{The dataset is divided into two subsets. The gold standard \ac{FID} is computed on the  two real subsets. The \ac{FID} of \ac{MDGAN} is computed based on one real subset and the generated copy of the other subset.}
    \label{fig:fid}
\end{figure}

\ac{FID}~\cite{fid} is a measure of the similarity between two sets of images (usually a set of real images and a set of \ac{GAN}-generated images). This metric was  used to assess the quality of the generated MR. The CT results are excluded from this analysis because of the much smaller dataset size, and they are also easier to translate compared to MR due to the lack of rich contrast information. Since there are no existing benchmark results on the MR dataset, we use the real data as the gold standard. As Figure~\ref{fig:fid}, we split the dataset $X$ into two subsets $X_1$ (1/3 of $X$) and $X_2$ (2/3 of $X$) and compute the gold standard \ac{FID} based on 50,000 slices sampled from each subset (with overlap within the subset). We then generate a fake replica of $X_1$ (denoted $X'_1$) using the segmentation maps of $X_1$, which also contains 50,000 slices (no overlap as output is non-deterministic). The \ac{FID} between $X_1$ and $X'_1$ are computed as \ac{MDGAN}'s performance metric. We take a 4-fold validation approach to this evaluation and the results are shown in Table~\ref{tbl:fid}. As a baseline, the \ac{FID} between the MR dataset and the CT dataset is 219.65. Therefore, our results are close to the golden standard with the margin of error.

\begin{table}[h]
    \caption{MDGAN FID results}
    \begin{tabular}{|l|llll|}
        \hline
                         & Split 1 & Split 2 & Split 3  & Split 4  \\ \hline
        Gold Standard MR & 22.77   & \textbf{15.78}   & \textbf{17.74}    & 33.47   \\
        MDGAN MR         & \textbf{20.30} & 19.56   & 22.71    & \textbf{14.81}    \\\hline
    \end{tabular}
    \label{tbl:fid}
\end{table}

\begin{figure*}[t!]
    \centering
    \includegraphics[width=\linewidth]{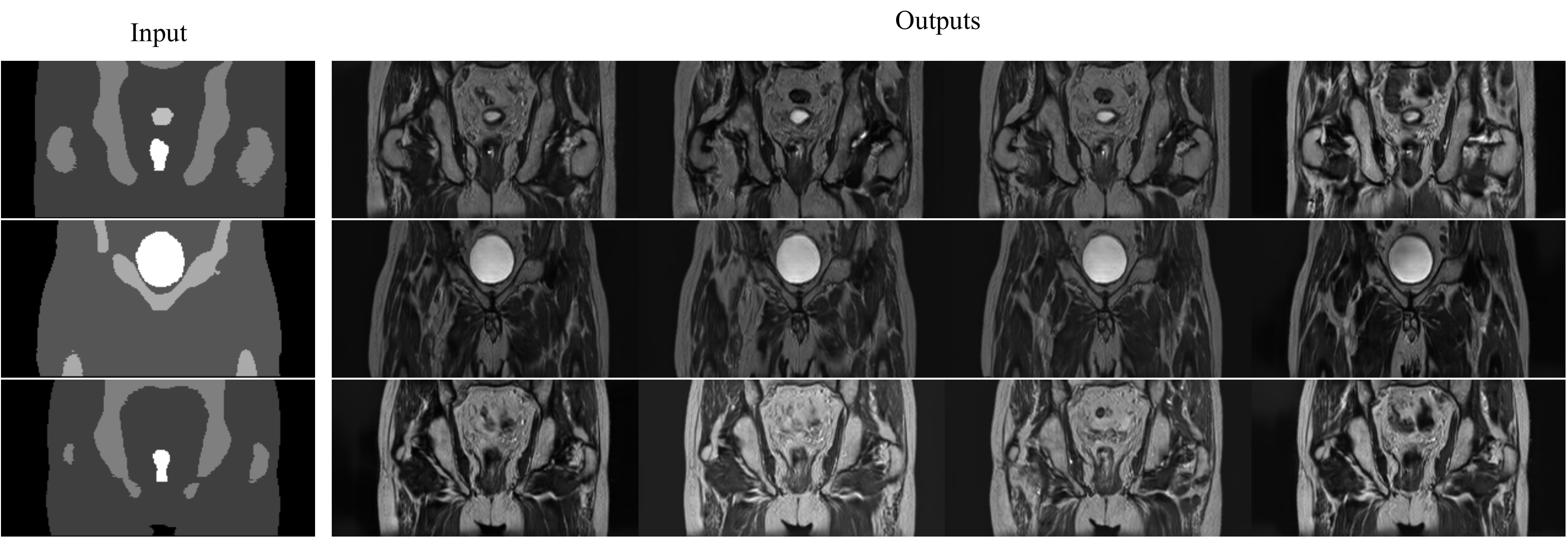}
    \caption{Diverse outputs generated using different style codes within the proposed \ac{MDGAN} from the same labelled image input.}
    \label{fig:results-div}
\end{figure*}

\begin{figure*}[t!]
    \centering
    \includegraphics[width=\linewidth]{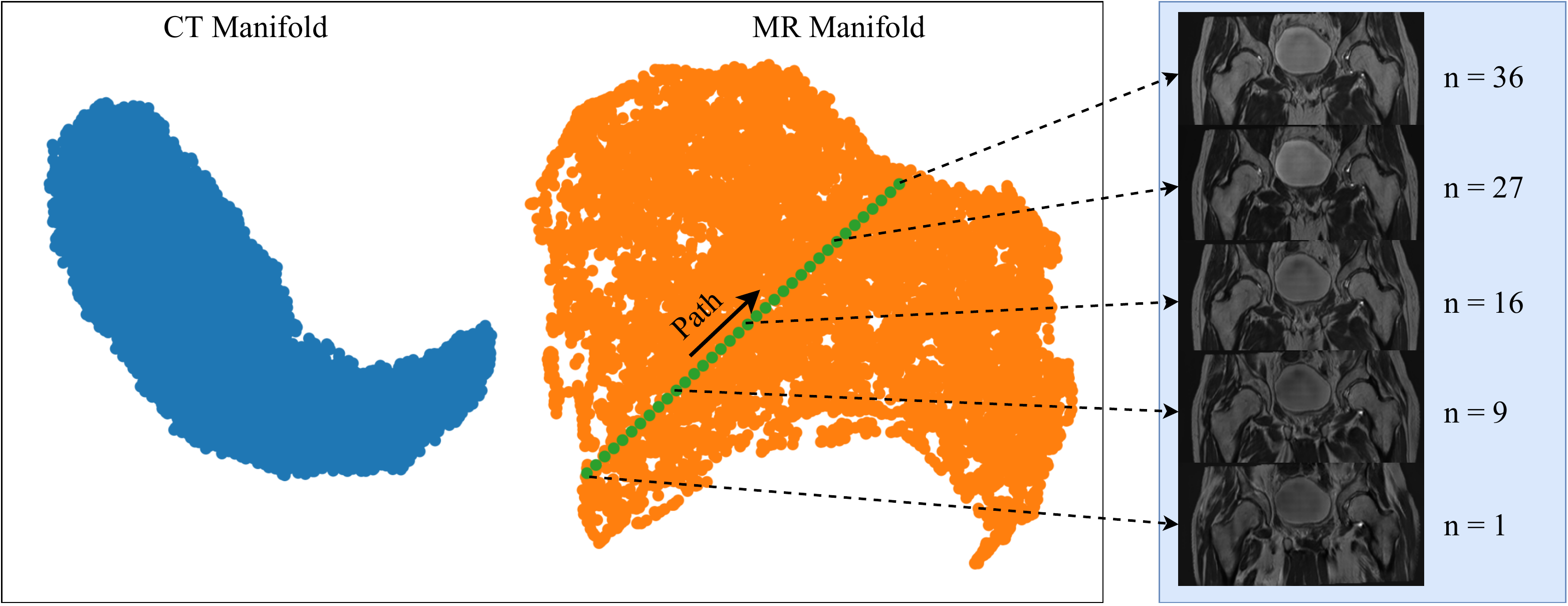}
    \caption{Left: 2D UMAP manifold mapping from 10,000 MR (orange) and 10,000CT (blue) style codes. The manifold clusters are naturally separated in 2D. The geodesic path chosen to explore the MR manifold is indicated in green. Right: images generated using the 1st, 9th, 16th, 27th and 36th points from the manifold path.}
    \label{fig:results-manifold}
\end{figure*}

\begin{figure*}[t!]
    \centering
    \includegraphics[width=\linewidth]{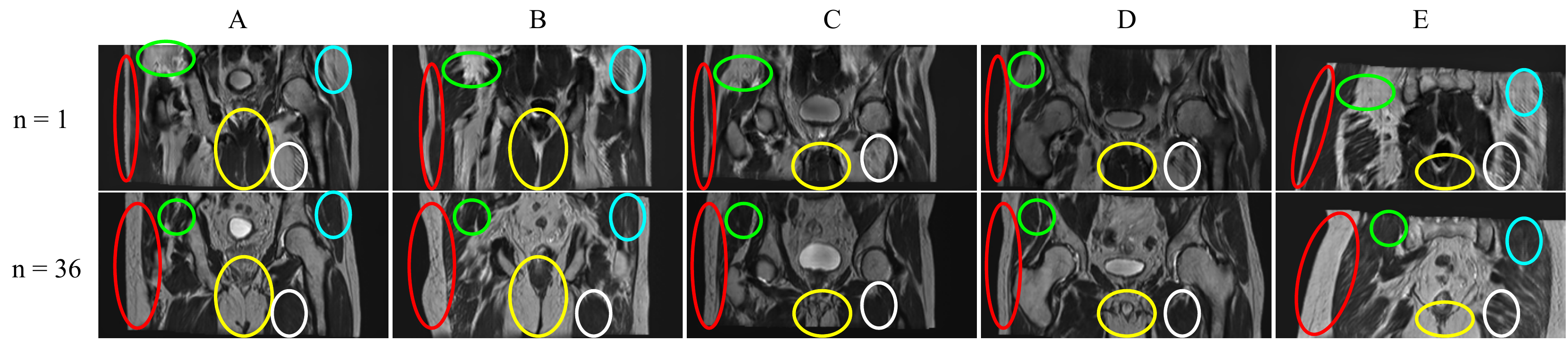}
    \caption{Change in tissue structures before (n=1) and after (n=36) transition.}
    \label{fig:results-manifold2}
\end{figure*}

\subsection{Diversity of Results}
To test \ac{MDGAN}~for output diversification, we perform image translation on a given input segmentation in combination with different style codes. The representative results in Figure~\ref{fig:results-div} suggest that the generator is well conditioned on both the segmentation map and the style code input. The explicit disentanglement of domain-invariant and domain-specific features allows us to ``edit'' the tissues in the generated MR while keep the mutual shape information intact. We also observe that scaling factor $\lambda_{div}$ has a positive association with the magnitude of diversification. Though large values of $\lambda_{div}$ takes significantly longer to converge and may occasionally produce invalid outputs.

\subsection{Manifold of MR Features}
Figure~\ref{fig:results-manifold} presents the results of the MR manifold geodesic path or ``walk''. As shown, the style codes of MR (orange) and CT (blue) are embedded as two separate manifold clusters on the manifold of the generator. The clear separation between the two clusters acts as a boundary to explicitly prevent ``feature mix ups" between the two exclusive domains. This suggests the proposed style-based generator is capable of learning and interpreting multiple medical imaging manifold clusters for different imaging modalities.

Traversing the chosen geodesic path in the MR manifold (green), the images generated using the style code sequence (and a chosen segmentation map) show smooth and systematic transitions. We observe that these transitions result into consistent and meaningful changes in tissue structure for all valid segmentation maps. Examples of this finding are shown in Figure~\ref{fig:results-manifold2} and the changes before and after transition are highlighted in colour. The style codes appear to have similar global influence on the all segmentation maps even for the heavily distorted case (Figure 7E) that was not part of the training set, though the features are localised differently in each image due to the shape and validity constraints.  $L_{\textnormal{div}}$ plays a profound role in forming these well-distributed manifold clusters. Our experiments without this term resulted in mode collapse, where all style codes produces the same output for a given input and chosen target-domain. Future work will involve using the proposed manifold disentanglement to construct meaningful manifolds in order to understand human diseases via MR and CT images from large studies.

Finally, all the images faithfully preserve the shape information as prescribed by the segmentation maps. Therefore, we believe the learnt manifold is disentangled from domain-invariant features, but semantically conditioned on the these features at the same time.

\section{Conclusion}
In this paper, we introduce \ac{MDGAN} as a style-based framework for medical image domain translation. Besides its robust generative performance, the frameworks explicitly models domain-invariant features and domain-specific features. We model domain-invariant features with a fully convolutional network, and domain-specific features as a disentangled manifold. We embed two manifold clusters onto the manifold using two style code manifold networks, which provide style codes for multi-modal (segmentation to MR and CT)  medical image translation. These manifold clusters are found to determine the target-domain as well as features specific to that domain in the image translation process. This valuable property could facilitate the detailed manifold learning of human diseases investigated with radiological techniques such as MR imaging.

\newpage
\bibliographystyle{IEEEtran}
\bibliography{ref}

\newpage
\section{Supplementary Experiments}
This material presents supplementary experiments to further explore the properties and potential applications of \ac{MDGAN}.

\subsection{Diverse MNIST Digit Generation from Labels}
\label{exp1}
This section presents a supplementary experiment using \ac{MDGAN} to generate hand-written digits~\cite{mnist} from labels. The goal is to test \ac{MDGAN} for more general conditional \ac{GAN} applications, where the input can be in abstract vectorised formats with no spatial information (in this case a one-hot encoded label vector). We show that the fundamental properties of manifold disentanglement discussed in the main paper still hold, and they can be exploited to meaningfully manipulate and diversify the output.

\begin{figure}[h!]
    \centering
    \includegraphics[width=\linewidth]{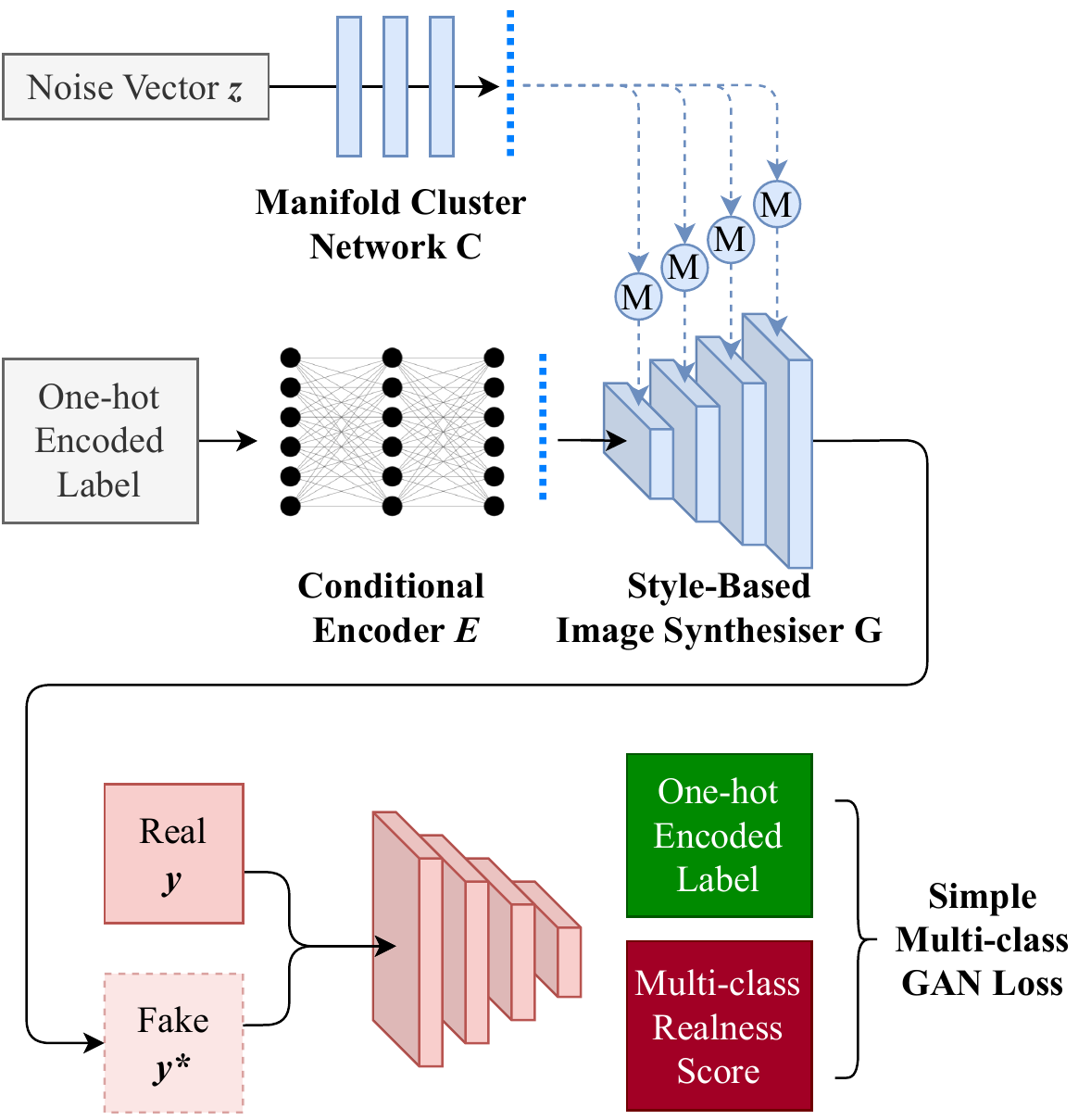}
    \caption{Modified \ac{MDGAN} architecture. The conditional encoder ($E$) is now a densely connected network. The discriminator is a CNN for multi-class realness assessment.}
    \label{fig:mnist1}
\end{figure}

% \subsubsection{Methods}
A heavily shrunk-down version of \ac{MDGAN} with minor modifications was used for this task. The modified architecture is shown in Figure~\ref{fig:mnist1}. The conditional encoder ($E$) now takes one-hot encoded labels (10-d) as the input, and encodes it via the conditional encoder ($E$), which is now a 3-layer densely connected network with 64 hidden units. The output from $E$ is then linearly projected onto a $7\times 7\times 64$-d vector space and reshaped into image representation format. Two up-sampling blocks (as seen in the main paper) with 64 and 32 filters are used to recover the original resolution. There is also a Manifold Cluster Network $C$ which provides style codes to manipulate the output without modifying the domain-invariant information (class label in this case). Finally, the discriminator $D$ is now multi-class and outputs a 10-d realness score (one for each class label, all zeros to represent fake). Like the generator, it is also shrunk down to include only two down-sampling blocks of 32 and 64 filters. The same loss functions were used without modification except the adversarial loss, which is now applied to every output class.

\begin{figure}[h]
    \centering
    \includegraphics[width=0.6\linewidth]{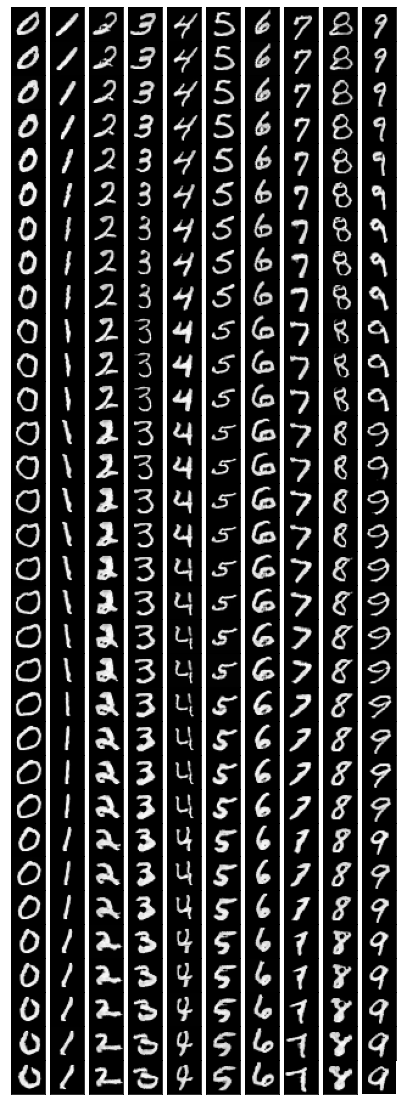}
    \caption{Image generated (for all 10 digits) along the path of traversal.}
    \label{fig:mnist-gen}
\end{figure}

\begin{figure}[h]
    \centering
    \includegraphics[width=0.6\linewidth]{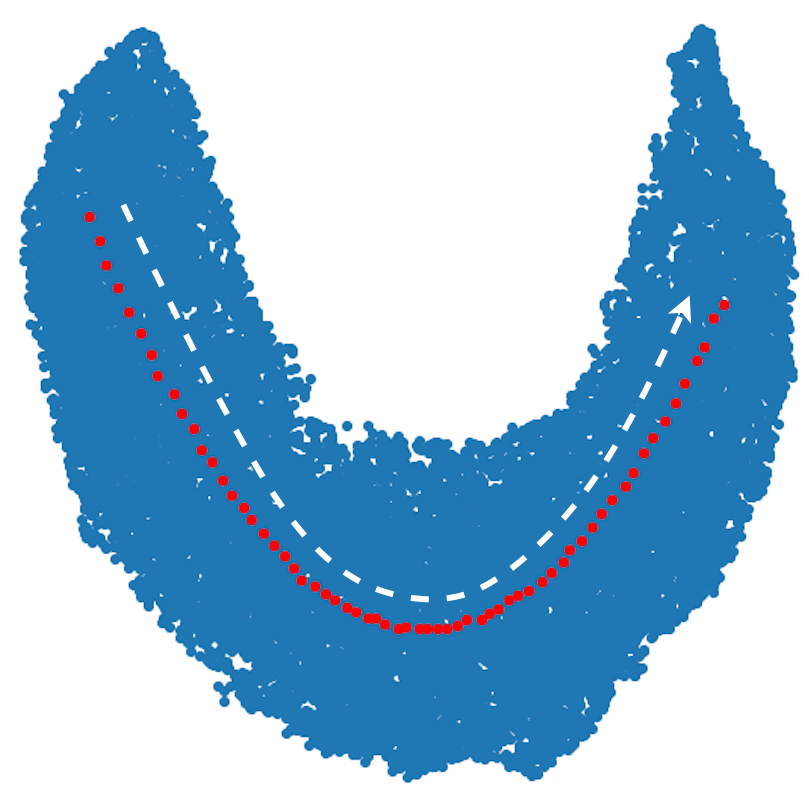}
    \caption{2-D visualisation of the manifold formation. The points along the chosen path of traversal are highlighted in red.}
    \label{fig:mnist-manifold}
\end{figure}

% \subsubsection{Results} 
Similar to section IV of the main paper, we use UMAP~\cite{umap} (with the same parameters) to perform dimensionality reduction on 10,000 sampled style codes. The manifold formation is visualised in 2D as Figure~\ref{fig:mnist-manifold}. A path (containing a series of points highlighted in red) is traversed to observe the transition of the generated image, and the generated images (for all 10 digits) are shown in Figure~\ref{fig:mnist-gen}. Consistent with the findings in the main paper, the learnt manifold of \ac{MDGAN} appears to be smooth as the generated images experience a smooth transition along the chosen path. The transition appears to be systematic. For example, the strokes appear thinner at the start for all the digits and become ``fuzzier" as the end. However, some of these trends can only be easily interpreted by the generator. As the disentanglement of the manifold is an important property enforced by \ac{MDGAN}, the input vector can be ``stylised" into diverse outputs of its class, but the validity and domain-invariant features (in this case their respective class label) are never violated. This shows \ac{MDGAN} can generalise beyond image-to-image translation tasks and can learn to construct manifold clusters based on other forms of conditional inputs.

\subsection{Diverse MNIST Digit Generation from Separate Manifold Clusters}
This section presents another supplementary experiment using \ac{MDGAN} to generate hand-written digits from labels. The goal is to provide further in-depth analysis on the localisation of spatial features (from the input) and domain-specific (manifold) features.

% \subsubsection{Methods}
Instead of starting $G$ with one-hot encoded labels, a $7 \times 7 \times64$ noisy image is used as the seed for $G$. We use the labels (0 to 9) to index one of the 10 separately learn Manifold Cluster Networks ($C_0, ..., C_9$) which determines the output class. The rest of $G$, the entirety of $D$ and the loss functions remain unchanged from Section~\ref{exp1}.

\begin{figure}[t]
    \centering
    \includegraphics[width=0.6\linewidth]{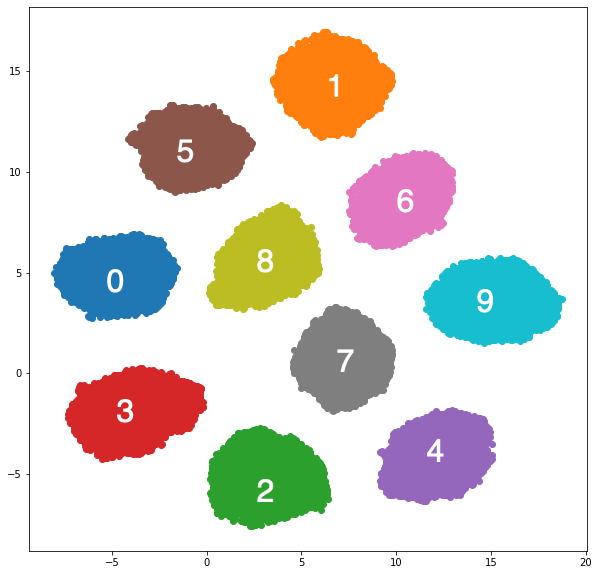}
    \caption{2-D visualisation of the manifold formation. The manifold clusters modelled $C_0, ..., C_9$ are shown as 10 labelled clusters.}
    \label{fig:mnist-manifold2}
\end{figure}

\begin{figure}[t]
    \centering
    \includegraphics[width=0.6\linewidth]{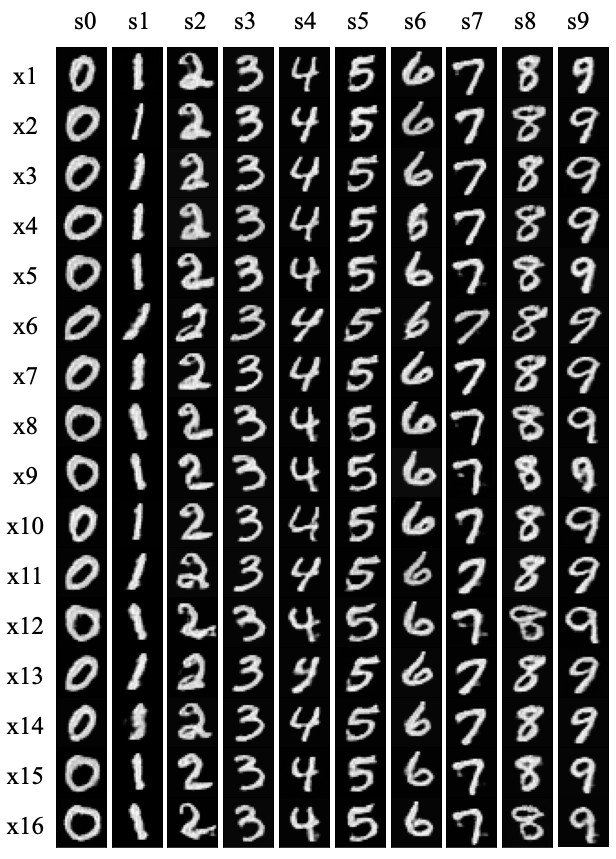}
    \caption{Images generated using 16 seeds ($x_1, ..., x_{16}$ and randomly-sampled style codes ($s_0, .., s_9$) from each class. The seed appears to determine properties such as ``font italicisation".}
    \label{fig:mnist-seed}
\end{figure}

\begin{figure}[t!]
    \centering
    \includegraphics[width=0.6\linewidth]{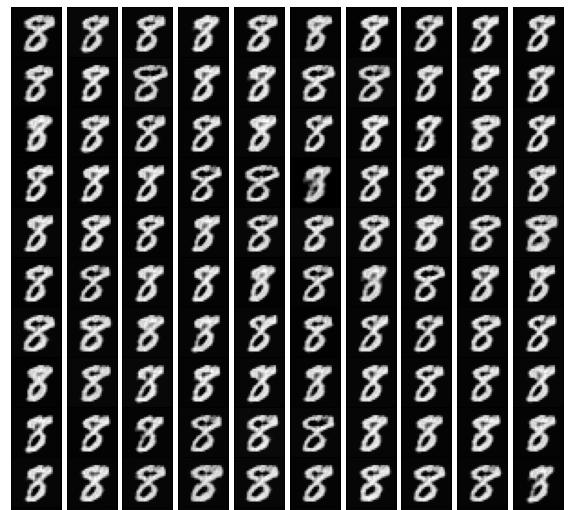}
    \caption{Images generated using a given seed $x$ and 100 style codes from $C_8$. All the images are different from each other, but the overall structure remains unchanged.}
    \label{fig:mnist-style}
\end{figure}

% \subsubsection{Results}
 5,000 style codes are sampled from each of $C_0, ..., C_9$, and they are mapped to 2-D (Figure~\ref{fig:mnist-manifold2}) using UMAP. As shown, the manifold clusters are distinctively separated. To investigate the localisation of domain-invariant features, 16 seeds ($x_1, ..., x_{16}$) and 10 style codes ($x_0, ..., x_9$, one from each class) are sampled to generate outputs and the results are shown in Figure~\ref{fig:mnist-seed}. It can be seen that while the style codes ultimately control the output class, the seeds determine more subtle domain-invariant properties such as ``font italicisation". For example, all the digits generated using $x_6$ are slanted to the right. As Figure~\ref{fig:mnist-style}, for a given seed, different style code sampled from a manifold cluster ($C_8$ in this case) can moderately alter the output while keeping the overall structure unchanged. This is because structural information is prescribed by the seed, which is a dense spatial input (less abstract compared to one-hot encoded labels) containing domain-invariant features. The style codes are thus localised as minor deviations from the main structure with relatively less ``freedom".

\acrodef{GAN}{\emph{Generative Adversarial Network}}
\acrodef{CNN}{\emph{Convolutional Neural Network}}
\acrodef{AdaIN}{\emph{Adaptive Instance Normalisation}}
\acrodef{MDGAN}{\emph{Manifold Disentanglement Generative Adversarial Network}}
\acrodef{FID}{Fréchet Inception Distance}
\end{document}